\documentclass[runningheads]{llncs}
\usepackage{amsmath,amsfonts,bm}

\def\eqref#1{equation~\ref{#1}}
\def\1{\bm{1}}

\DeclareMathAlphabet{\mathsfit}{\encodingdefault}{\sfdefault}{m}{sl}
\SetMathAlphabet{\mathsfit}{bold}{\encodingdefault}{\sfdefault}{bx}{n}

\usepackage{ulem}
\usepackage{soul}
\usepackage{url}
\usepackage{booktabs}       % professional-quality tables
\usepackage{amsfonts}       % blackboard math symbols
\usepackage{nicefrac}       % compact symbols for 1/2, etc.
\usepackage{bbding}
\usepackage{amssymb}
\usepackage{multirow}
\usepackage{caption}
\usepackage{subcaption}
\usepackage{graphicx}
\usepackage{xcolor}
\usepackage{float}
\usepackage{colortbl}
\usepackage{listings}
\usepackage{makecell}
\usepackage{wrapfig}
\usepackage{cellspace}
\usepackage[misc]{ifsym}
\usepackage[utf8]{inputenc}

\definecolor{lz}{rgb}{.224,.451,.686}

\usepackage{microtype}
\usepackage{inconsolata}
\newcommand{\amlpone}{AMLP-Cov}
\newcommand{\amlptwo}{AMLP-PQuery}
\usepackage{graphicx}

\begin{document}

\title{Attentive Multi-Layer Perceptron\\ for Non-autoregressive Generation}
%
% \titlerunning{AMLP}
% \orcidID{0009-0006-7423-6333}
\author{Shuyang Jiang\inst{1} \and Jun Zhang\inst{2} \and Jiangtao Feng\inst{3} \and Lin Zheng\inst{4} \and Lingpeng Kong\inst{4}$^{(\text{\Letter})}$}
\authorrunning{S. Jiang et al.}
\institute{Shanghai Jiao Tong University, Shanghai, China \\
\email{jiangshuyang@sjtu.edu.cn} \and
Shanghai Artificial Intelligence Laboratory, Shanghai, China \\
\email{zhangjun@pjlab.org.cn}  \and 
\email{jiangtaofeng0906@gmail.com}
\and 
The University of Hong Kong, Hong Kong, China \\ 
\email{\{linzheng@connect,lpk@cs\}.hku.hk} \\
% \email{lpk@cs.hku.hk}
}

\maketitle              % typeset the header of the contribution

\begin{abstract}
Autoregressive~(AR) generation almost dominates sequence generation for its efficacy.
Recently, non-autoregressive~(NAR) generation gains increasing popularity for its efficiency and growing efficacy.
However, its efficiency is still bottlenecked by quadratic complexity in sequence lengths, which is prohibitive for scaling to long sequence generation and few works have been done to mitigate this problem.
In this paper, we propose a novel MLP variant, \textbf{A}ttentive \textbf{M}ulti-\textbf{L}ayer \textbf{P}erceptron~(AMLP), to produce a generation model with linear time and space complexity.
Different from classic MLP with static and learnable projection matrices, AMLP leverages adaptive projections computed from inputs in an attentive mode.
The sample-aware adaptive projections enable communications among tokens in a sequence, and model the measurement between the query and key space.
Furthermore, we marry AMLP with popular NAR models, deriving a highly efficient NAR-AMLP architecture with linear time and space complexity.
Empirical results show that such marriage architecture surpasses competitive efficient NAR models, by a significant margin on text-to-speech synthesis and machine translation.
We also test AMLP's self- and cross-attention ability separately with extensive ablation experiments, and find them comparable or even superior to the other efficient models.
The efficiency analysis further shows that AMLP extremely reduces the memory cost against vanilla non-autoregressive models for long sequences.

\end{abstract}

\section{Introduction}
\label{intro}
Attention-based sequence generation methods have achieved great success and gained increasing popularity in machine learning~\cite{vaswani2017attention,transformer_tts,liu2021swin,dosovitskiy2020image}.
A large body of research in neural architectures has been devoted to the autoregressive (AR) method~\cite{abc,RFA}, where tokens are generated one after another in an iterative manner.
The computational overhead in decoding can thus be prohibitive, especially for long sequences.
Recently, non-autoregressive (NAR) generation attracts more attention for its efficiency and growing efficacy~\cite{gu2017non,gu2019levenshtein,qian2020glancing,qian2021volctrans,fastspeech2,chang2022maskgit}.
In a non-autoregressive model, the decoder generates the target sequence all at once, significantly reducing its computational overhead at the inference stage.
Nevertheless, relatively little research has been done on the attention architecture in non-autoregressive models. In particular, the conventionally adopted softmax attention comes with a quadratic time and memory cost. It is therefore still difficult to scale up non-autoregressive models to long sequence generation tasks.

In this paper, we propose Attentive Multi-Layer Perceptron (\S\ref{amlp_arch}; AMLP) to integrate the attention mechanism with the multi-layer perceptron (MLP) in non-autoregressive architecture, resulting in a fully parallelizable sequence generation model with linear complexity.
Unlike the widely-used MLP whose weights are invariant across different sequences, we compute the weights in AMLP through adaptive projections from (multiple) input tokens and model their interactions in an attentive manner.
Specifically, we put forward two methods~(\S\ref{sec:parameterize_weights}) to compute the adaptive projections in AMLP, which implicitly model the association between the query and key space. 
We utilize the simplicity and efficiency of MLP while obtaining the strong modeling capability of AMLP for input tokens' communication.
Finally, we present a hybrid NAR-AMLP model~(\S\ref{sec:hybrid_arch}) to achieve both linear complexity and high parallelism.

We evaluate the AMLP architecture on text-to-speech synthesis for a relatively long sequence scenario and machine translation for a relatively short sequence scenario. 
Experiments show that AMLP achieves more superior scores with objective measurements compared with the strong softmax attention counterpart (\S\ref{sec:full_linear_experiment}) on text-to-speech synthesis, with less computational cost (\S\ref{ablation}). 
On machine translation, AMLP performs competitive with vanilla attention but achieves the best result among efficient NAR and AR models with linear complexity~(\S\ref{sec:mt}).
Further, we test the self- and cross-attention ability of AMLP on super resolution and long sequence time-series forecasting tasks, respectively.
Empirical results show that AMLP is on par with other efficient attention in self-attention and achieves the best performance in cross-attention scenarios~(\S\ref{sec:ablate_self_cross}).
Additionally, when scaling to long sequence, AMLP reduces the memory footprint substantially and further improves the inference speed in NAR models~(\S\ref{efficiency_analysis}).
The code is available in \url{https://github.com/Shark-NLP/AttentiveMLP}.

\begin{figure}[tbp]
    \centering
    \begin{subfigure}[c]{0.48\textwidth}
         \centering
         \includegraphics[width=\textwidth]{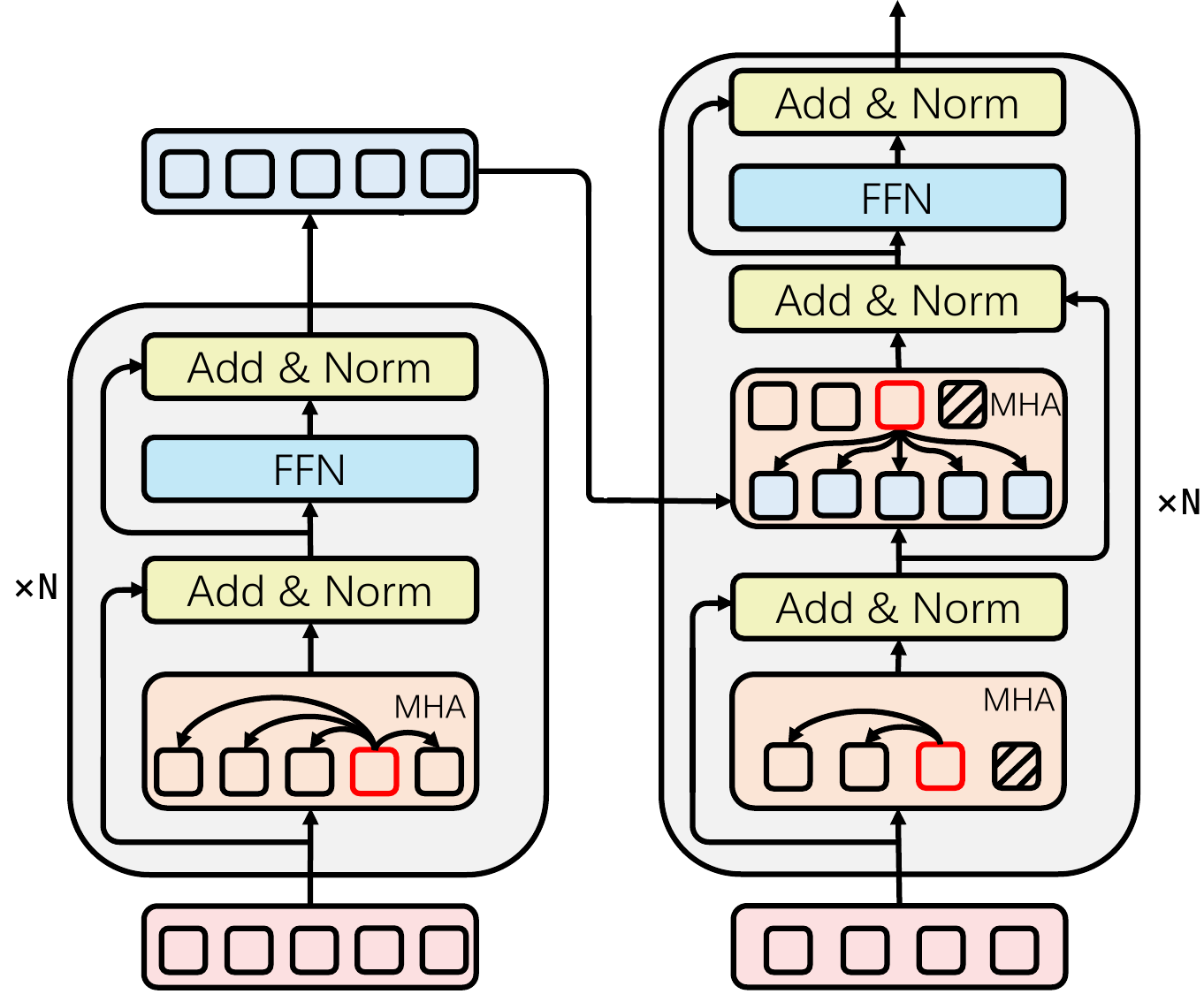}
         \caption{%Empirical test on running time. These results are time consumption of 100 unit tests on average.
         }
         \label{fig:ar}
     \end{subfigure}
     % \hspace{\fill}
     \begin{subfigure}[c]{0.48\textwidth}
         \centering
         \includegraphics[width=\textwidth]{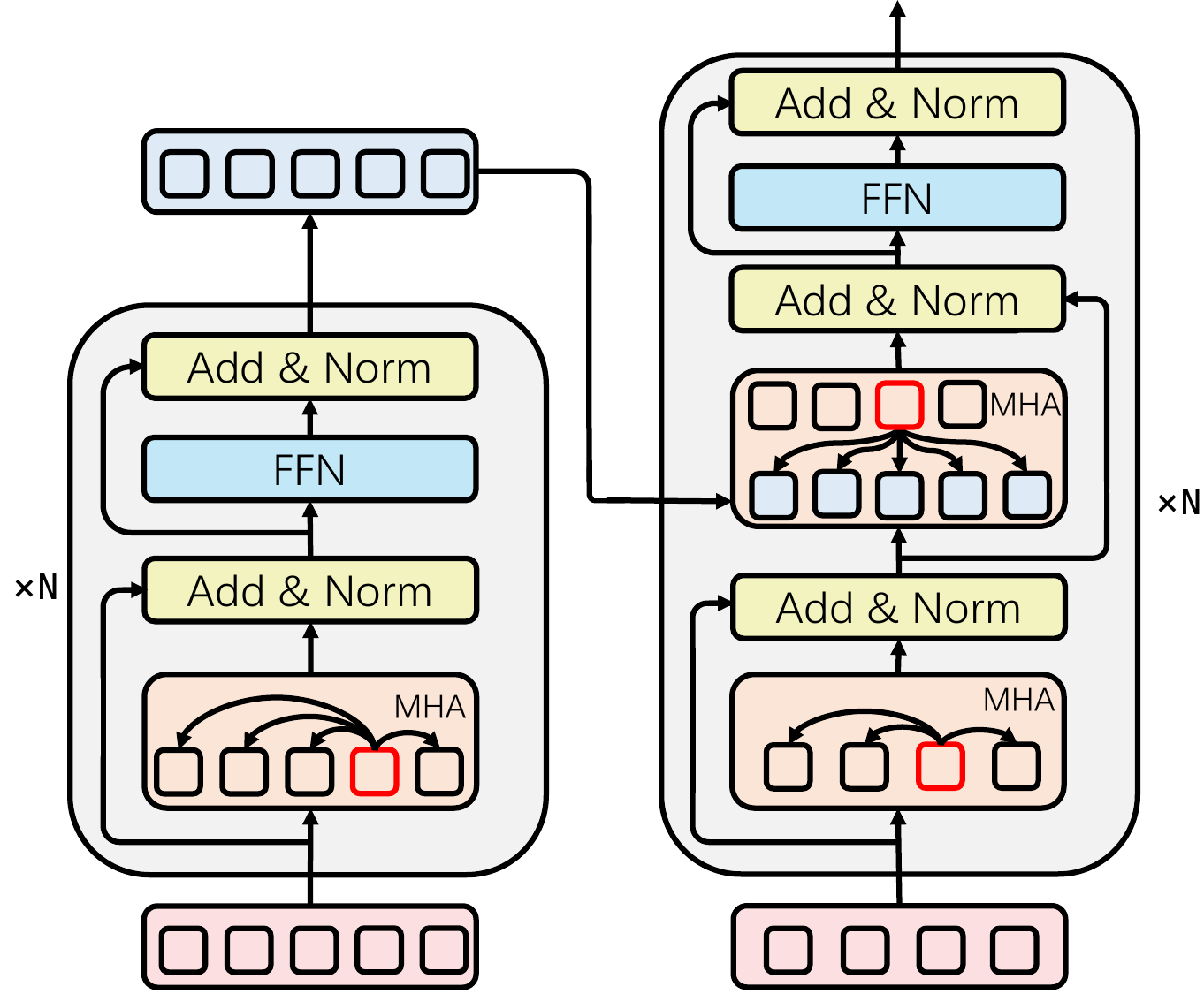}
         \caption{%Empirical test on peak GPU usage. These results are peak memory consumption during the process of computation.
         }
         \label{fig:nar}
     \end{subfigure}
    \caption{
    AR (a) and NAR (b) encoder-decoder architectures. ``MHA'' stands for multi-head attention. Blocks with red rims represent the current state token. Shaded blocks represent future tokens that are invisible to the current state. 
    }
    \label{fig:ar_nar_taxonomy}
\end{figure}

\section{Non-Autoregressive Generation with Attentive MLP}

In this section, we first give a brief introduction to autoregressive~(AR) and non-autoregressive~(NAR) generation, and then delve into the nuances that differentiate the attention mechanisms utilized in autoregressive (AR) and non-autoregressive (NAR) models. 
After that, we present the AMLP architecture to model the communication among sequence tokens.
Finally, we build up an NAR-AMLP architecture with linear time and space complexity.

\subsection{Background: Autoregressive and Non-Autoregressive Generation}
Given a source sequence $X_{1:m}$, conditional sequence generation targets to predict a target sequence $Y_{1:n}$ by modeling the conditional probability $p(Y|X)$.
Autoregressive generation decomposes the probability $p(Y|X)$ as:
\begin{equation}
    p(Y|X)=\prod_{i=1..n} p(Y_i|Y_{<i}, X), Y_{<1}=\emptyset.
\end{equation}
which is implemented as a typical encoder-decoder architecture shown in Fig.~\ref{fig:ar}.
Although such decomposition is proved effective, it suffers from two main drawbacks: efficiency and exposure bias.
On the one hand, the autoregressive decoding process, where each token depends on the previous predicted ones, prevents the model from fast inference in usage.
On the other hand, teacher-forcing exposes ground truth tokens in network inputs during the training process, where the exposed tokens are unable to observe in inference.
Such exposure creates an inconsistency between the training and inference, and harms the prediction quality. 

Recently, non-autoregressive generation, depicted as Fig.~\ref{fig:nar}, shows its capability of sequence modeling in terms of both efficiency and efficacy, which decomposes the conditional probability $p(Y|X)$ via a Na\"ive Bayes assumption:
\begin{equation}
    p(Y|X)=\prod_{i=1..n} p(Y_i|X)
\end{equation}
The NAR decomposition enables parallel decoding for each token, and speeds up the
inference process substantially.
Although NAR generation is much faster than AR generation, its speed is still limited by the $O\left(n^2+nm+m^2\right)$ time complexity of the multi-head softmax attention module. This is especially problematic in modeling long sequences.

\subsubsection{Attention Types in AR \& NAR Models}
Although autoregressive and non-autoregressive models differ from each other in sequence generation paradigms, their underlying attention mechanisms in their architectures are also different.
The token-by-token generation of AR models requires a causal decoder that forces tokens to attend to only previous features.
A typical causal decoder utilizes causal softmax attentions both in self-attention and cross-attention.
The attention causality entails that during the computation, it is important to ensure that the query token does not attend to the context on its right side, just as the shaded blocks in Fig.~\ref{fig:ar}. 
In contrast, the NAR model, which allows for parallel generation of the output sequence and global contextualization using attention, employs a noncausal decoder in Fig.~\ref{fig:nar}.
The self-attention in the NAR model can attend to both side contexts of a given token, which makes it suitable for tasks that require a broader contextual understanding. 
NAR architectures also reduce the design restrictions on cross-attention, making query tokens attend to key tokens in a holistic view.
This modeling feature of attention emphasizes both global and local contextualization modeling for attention modules.
In practice, causality in vanilla softmax self-attention is ensured by leveraging a lower triangular mask in AR models, while linearized attention requires more sophisticated implementation.
Since no causality is required in NAR models, designing an efficient attention mechanism is much more flexible.

\subsection{Attentive Multi-Layer Perceptron}
\label{amlp_arch}
Modeling interactions between tokens is crucial and challenging in sequence generation.
Transformer~\cite{vaswani2017attention} stacks the MLP, which aims to learn features of individual tokens, on top of the attention block, which is responsible for modeling the communication within the sequence. 
In AR generation, the attention needs to be recomputed for each time step through the recurrent process, as the key and value set is changing. However, this procedure is non-causal in NAR generation.
We therefore are able to integrate the modeling of token interactions into the MLP architecture and make the whole architecture fully parallelizable and more efficient.

Given a sequence representation $\mathbf{X}\in\mathbb{R}^{n\times d}$, where $n$ is the sequence length and $d$ is dimensionality of the feature space, the conventional MLP models the feature of individual token $\mathbf{X}_i\in\mathbb{R}^{d}$ as:
\begin{equation}
\label{eq:mlp}
    \text{MLP}(\mathbf{X}_i) = \sigma(\mathbf{X}_i W_1)W_2
\end{equation}
where $W_1\in\mathbb{R}^{d\times d_h}$, $W_2\in\mathbb{R}^{d_h\times d}$ are learnable parameters $d_h$ is the dimensionality of hidden space.
$\sigma(\cdot)$ is a non-linear activation function such as $\mathrm{ReLU}(\cdot)$.
However, it disables the communication between tokens in the sequence, and prevents the model from learning contextualized token representations.

A widely-used approach to enable communication between each token in a sequence is the attention mechanism~\cite{vaswani2017attention}.
Vanilla attention learns to incorporate source sequence features $\mathbf{K,V}\in\mathbb{R}^{m\times d}$ into target $\mathbf{Q}\in\mathbb{R}^{n\times d}$ with an attention matrix
\begin{equation}
    \label{eq:attention}
    \mathrm{Attn}(\mathbf{Q,K,V})=\mathrm{softmax}(\mathbf{QK^\top})\mathbf{V}
\end{equation}
where $m$, $n$ are the source and target length respectively.
Here we omit the input projections for $\mathbf{Q,K,V}$, the output projection, and the scaling factor~$1/\sqrt{d}$ for simplicity.

The motivation of Attentive Multi-Layer Perceptron~(AMLP) starts from the fact that the vanilla softmax attention can be viewed as a projection function as $\mathrm{SA}(\cdot | \mathbf{K, V}): \mathbb{R}^{n \times d} \rightarrow \mathbb{R}^{n \times d}$ which projects the original $\mathbf{Q} \in \mathbb{R}^{n \times d}$ with $\mathbf{K}$ and $\mathbf{V}$ features as its context while preserving $\mathbf{Q}$'s shape. 
In vanilla attention, $\mathrm{softmax}(\mathbf{QK}^\top)$ is a softmax kernel which can be decomposed into a multiplication of two kernel functions: $\phi(\mathbf{Q})\cdot\phi(\mathbf{K})^\top$, which is verified in Performer~\cite{choromanski2020rethinking}, cosFormer~\cite{qin2022cosformer} and LARA~\cite{lara}. 
Meanwhile, the low-rank factorization of the attention matrix, $\mathrm{softmax}(\mathbf{QK}^\top)$,  does not impact the performance much, which is verified by Nystr\"omformer~\cite{xiong2021nystromformer}. 
Based on their findings, we propose an alternative modeling solution by fusing key $\mathbf{K}\in \mathbb{R}^{m\times d}$ and value $\mathbf{V}\in\mathbb{R}^{m\times d}$ information into query $\mathbf{Q}\in \mathbb{R}^{n\times d}$, via a symmetric and positive semi-definite distance matrix $\mathbf{\Sigma}\in\mathbb{R}^{d\times d}$ on $\mathbf{Q}$ and $\mathbf{K}$ space. 
The contextualizing process on $\mathbf{Q}$ can be formulated as:
\begin{equation}
\label{eq:psd}
    f(\mathbf{Q;K, V}) = \mathbf{Q} \mathbf{\Sigma} \mathbf{K^\top V}
\end{equation}
where $\mathbf{\Sigma}$ is computed from $\mathbf{Q}$ and $\mathbf{K}$.

With similar functionality to~\cite{choromanski2020rethinking,xiong2021nystromformer}, the matrix $\mathbf{Q} \mathbf{\Sigma} \mathbf{K}^\top$ can also enjoy lower computation costs from low-rank approximation while maintaining strong modeling capability.
Without taking any low-rank assumptions on input $\mathbf{Q,K}$, we decompose the distance matrix as:
\begin{equation}
    \label{ea:decom_sigma}
        \mathbf{\Sigma} = \mathbf{U\Lambda U^\top} = \mathbf{U\Lambda ^{\frac{1}{2}}\Lambda^{\frac{1}{2}}U^\top} \approx \mathbf{U\hat{\Lambda} ^{\frac{1}{2}}\hat{\Lambda}^{\frac{1}{2}}U^\top} \\
        =\mathbf{(U\hat{\Lambda}^{\frac{1}{2}})(U\hat{\Lambda}^{\frac{1}{2}})^\top}= \mathbf{LL^\top}
\end{equation}
where $\mathbf{U}$ is the orthogonal eigenvector of matrix and $\mathbf{\Lambda}$ is the diagonal eigenvalues matrix.
$\hat{\mathbf{\Lambda}}$ here is an approximation to $\mathbf{\Lambda}$ by keeping largest-$c$ eigen-values and masking the others with $0$, where $c$ is a hyper-parameter in AMLP.
Thus we derive a decomposition equation $\mathbf{\Sigma} \approx \mathbf{LL^\top}$ where $\mathbf{L}=\kappa(\mathbf{Q,K})^\top\in \mathbb{R}^{d\times c}$ indicates a low-rank matrix.
We will show two different methods for parameterization of $\mathbf{L}$, resulting in two different AMLP variants.
We rewrite Eq.~\ref{eq:psd} by decomposing the distance matrix $\mathbf{\Sigma}$ as:
\begin{equation}
\label{eq:decomp}
    f(\mathbf{Q, K, V}) \approx \mathbf{QLL^\top K^\top V} 
\end{equation}
Now Eq.~\ref{eq:psd} could be approximated with Eq.~\ref{eq:decomp} by linearly projecting the original $\mathbf{Q}$ with adaptive weights twice. 
By reordering the computation and adding nonlinearity into Eq.~\ref{eq:psd}, we derive a general form of AMLP model as:
\begin{equation}
\label{eq:nonlinear}
    \mathrm{AMLP}(\mathbf{Q;K, V})=\sigma_1(\mathbf{Q}W_{\mathbf{Q},\mathbf{K}})W_{\mathbf{Q},\mathbf{K},\mathbf{V}}
\end{equation}
where the nonlinear function $\sigma_1(\cdot)$ can be adjusted arbitrarily.
Eq.~\ref{eq:nonlinear} address the general form of AMLP, and the adaptive weights $W_{\mathbf{Q},\mathbf{K}}$ and $W_{\mathbf{Q},\mathbf{K},\mathbf{V}}$ can be speficified in various ways.
% We show two of their parameterization in \S~\ref{sec:parameterize_weights}.
Following the form of Eq.~\ref{eq:nonlinear}, we will further introduce two AMLP variants in~\S~\ref{sec:parameterize_weights}, by specifying $\mathbf{L}=W_{\mathbf{Q,K}}=\kappa(\mathbf{Q},\mathbf{K})$, computational order and nonlinear function.

The computation of adaptive weights in AMLP fuses token-level communication, while MLP models tokens in a sequence independently. 
Therefore, AMLP enables the communication between tokens in a sequence.
And different from vanilla softmax attention, AMLP utilizes a distance matrix $\mathbf{\Sigma}$ between $\mathbf{Q}$ and $\mathbf{K}$ spaces to fuse information among their contexts and outputs a contextualized $\mathbf{Q}$.
Through this distance matrix, AMLP computes the similarity between $\mathbf{Q}$ and $\mathbf{K}$ like softmax attention, and leverages it to aggregate $\mathbf{V}$.

\begin{figure*}[tbp]
    \centering
    \includegraphics[width=\textwidth]{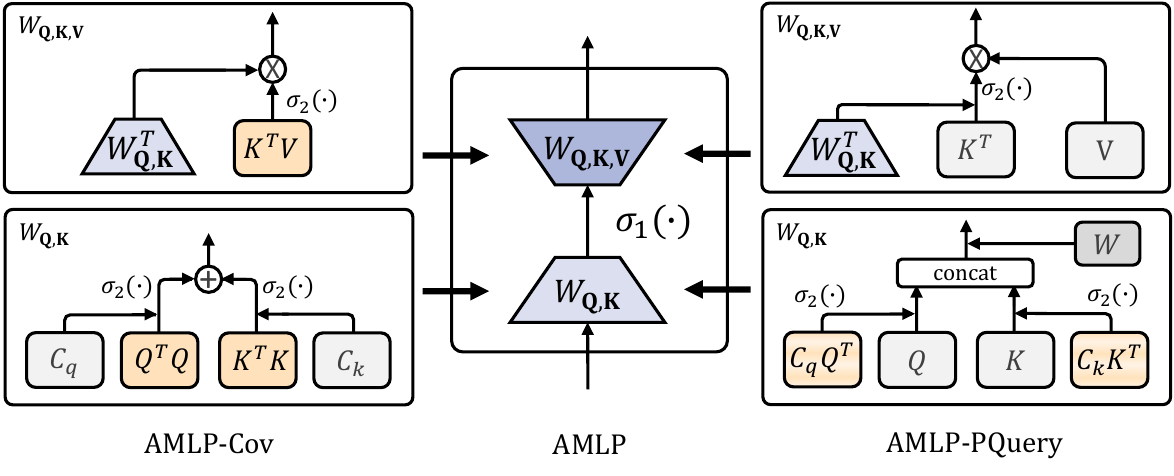}
    \caption{Computation diagram of two AMLP variants. The middle part shows the computation of basic AMLP. The Left and right figures show the detailed computation of two adaptive weight matrics in \amlpone~and \amlptwo. }
    \label{fig:nar_fully_mlp}
\end{figure*}
\subsection{Parameterization}
\label{sec:parameterize_weights}
In this section, we describe two methods for the parameterization of two adaptive weight matrices $W_{\mathbf{Q,K}}$ and $W_{\mathbf{Q,K,V}}$. 
Fig.~\ref{fig:nar_fully_mlp} illustrates the computation graph of these two methods.~\footnote{AMLP is implemented with multiple heads~\cite{vaswani2017attention}, but for simplicity and without loss of generality, we will discuss our AMLP computation process in a single-head setting.}

\subsubsection{Cross-Covariance}
We present \amlpone, a variant that adopts cross-covariance to parameterize $W_{\mathbf{Q,K}}$ and $W_{\mathbf{Q,K,V}}$.
One challenge of AMLP is to fuse information of $\mathbf{Q,K,V}$ of different shapes into static-shaped projection matrices $W_{\mathbf{Q,K}}$ and $W_{\mathbf{Q,K,V}}$.
Inspired by~\cite{el2021xcit}, we propose to use $\mathbf{Q,K}$'s covariance and the cross-covariance between $\mathbf{K}$ and $\mathbf{V}$ in AMLP.
To obtain $\mathbf{L}=\kappa(\mathbf{Q,K})^\top$, we separately compute $\mathbf{Q}$'s and $\mathbf{K}$'s covariance matrices and combines them with learned down-sampling projection matrices ${C}_q \in \mathbb{R}^{c \times d}$ and ${C}_k \in \mathbb{R}^{c \times d}$:

\begin{equation}
\label{eq:weightqk}
    \mathbf{\kappa(Q, K)}={C}_q\left(\sigma_2\mathbf{(Q^\top Q)}\right)+{C}_k\left(\sigma_2\mathbf{(K^\top K)}\right)
\end{equation} 
where $\sigma_2(\cdot)$ is set to softmax function as~\cite{el2021xcit} suggest.
The covariance matrices of $\mathbf{Q,K}$ are of the same shape and can be directly fused.
We add the softmax function as a non-linear activation to enhance the expressiveness.
For $W_{\mathbf{Q,K,V}}$, we notice the shapes of $\mathbf{K}$ and $\mathbf{V}$ are usually identical, and we hence use their cross-covariance $\mathbf{K}^\top \mathbf{V}$ for computation in Eq.~\ref{eq:nonlinear}.
$W_{\mathbf{Q,K,V}}$ is then formulated by transforming the cross-covariance $\mathbf{K}^\top \mathbf{V}$ to query space by $\mathbf{L}$ as:
\begin{equation}
    \label{eq: weight_qkv}
    W_{\mathbf{Q,K,V}}=\mathbf{L}^\top\sigma_2(\mathbf{K}^\top \mathbf{V})
\end{equation}

\subsubsection{Pseudo-Queries}
\amlptwo~first uses Exponential Moving Average (EMA) to compute the contextualized query via a hyperparameter $\beta$: $\mathbf{\hat{q}_i}=\beta\cdot \mathbf{\hat{q}}_{i-1}+(1-\beta)\cdot \mathbf{q}_i$, which has been proved to model local context well~\cite{ma2022mega}.
To further improve the communication between target and source sequences in a long sequence view, \amlptwo~treats learnable $C_q$, $C_k$ and $\mathbf{L}^\top$ as pseudo attention queries.
Specifically, it estimates $W_{\mathbf{Q,K}}$ by fusing features from query and key to the hidden space with an extra learnable weight $W\in\mathbb{R}^{2d\times d}$:
\begin{equation}
    \label{eq:amlpseq_qk}
    W_{\mathbf{Q,K}}=\mathbf{L}^\top=\left[\sigma_2({C}_q\mathbf{\hat{Q}^\top) \hat{Q}};\sigma_2({C}_k\mathbf{K^\top) K}\right]W
\end{equation} 
where $\sigma_2(\cdot)$ is set to softmax as \amlpone.
For $W_{\mathbf{Q,K,V}}$, we notice that $\mathbf{L}^\top$ has fused features from $\mathbf{\hat{Q}}$.
So we again treat $\mathbf{L}^\top$ as a pseudo query to fuse features from the source sequence:
\begin{equation}
\label{eq:amlpseq_qkv}
    W_{\mathbf{Q,K,V}}=\sigma_2(\mathbf{L^\top K^\top) V}
\end{equation}
With explicit communication between $\mathbf{\hat{Q}}$ and $\mathbf{K}$ in $W_{\mathbf{Q,K,V}}$, the alignment between different sequences is enhanced; therefore, \amlptwo~is more adaptive to cross-attention.

\subsection{Linear NAR: A Hybrid Architecture of NAR and AMLP}
\label{sec:hybrid_arch}
We combine AMLP with NAR for lower memory costs, faster inference speed and higher parallelism because AMLP and NAR are mutually reinforcing.
% why nar needs amlp
\subsubsection{AMLP boosts NAR}
On one hand, NAR parallelizes the inference process, but its efficiency is still hindered by vanilla attention. 
AMLP, as a plug-in efficient attentive module, mitigates the inefficiency effortlessly.
On the other hand, the non-autoregressive pipeline provides a non-causal encoding framework, with which the computation of AMLP avoids fine-grained operations.
\subsubsection{NAR augments AMLP}
We present the specific computation steps of AMLP in AR scenario and explain the drawbacks of AR-AMLP.
We take \amlpone~as an example.
Given an query token $\boldsymbol{q}_t$, the covariances $\mathbf{S}_t^{\mathbf{Q}}$ and $\mathbf{S}_t^{\mathbf{K}}$ of $\mathbf{K}_{t}$ and $\mathbf{Q}_{t}$, and the cross-covariance $\mathbf{z}_t$ of $\mathbf{K}_t$ and $\mathbf{V}_t$, $W_{\mathbf{Q,K}}$ and $W_\mathbf{{\mathbf{Q,K,V}}}$ are formulated as:
\begin{align}
    \label{eq:causal_weight}
    W_{\mathbf{Q}_{t},\mathbf{K}_t}=\mathbf{L}_t^\top&=C_q(\sigma_2(\mathbf{S}_t^{\mathbf{Q}}))+C_k(\sigma_2(\mathbf{S}_t^{\mathbf{K}})) \\
    W_{\mathbf{Q}_{t},\mathbf{K}_t, \mathbf{V}_t}&=\mathbf{L}_t^\top\sigma_2(\mathbf{z}_t)
\end{align}
where $\mathbf{S}_t^{\mathbf{Q}}=\mathbf{S}_{t-1}^{\mathbf{Q}}+\boldsymbol{q}_t^\top\boldsymbol{q}_t$, $\mathbf{S}_t^{\mathbf{K}}=\mathbf{S}_{t-1}^{\mathbf{K}}+\boldsymbol{k}_t^\top\boldsymbol{k}_t$ and $\mathbf{z}_t=\mathbf{z}_{t-1}+\boldsymbol{k}_t^\top\boldsymbol{v}_t$.
These computation steps increase heavy memory costs and large time consumption in the training phase, with an additional $O(ncd)$ costs beyond the overall computation.
Recurrent computation also harms the parallelism and further slows down the training process, which is avoided naturally in NAR models.
Moreover, CAB~\cite{zhang2022cab} points out that most existing efficient architectures suffer a great performance drop in causal-self or causal-cross pattern of AR models.
Combining the two drawbacks brought by the fusion of efficient architecture and AR models, we decide to incorporate AMLP into NAR to produce a powerful and efficient model.

\subsection{Complexity Analysis}
Without loss of generality, we focus on the complexity in the typical encoder-decoder architecture and omit the independent factor \textit{w.r.t.} target length $n$ and source length $m$ for simplicity.
\subsubsection{\amlpone~\& \amlptwo}
Note that the inner dimension $c$ is a constant to both $m$ and $n$.
The sequential computation of two adaptive projection matrices and the overall MLP computation  in Eq.~\ref{eq:nonlinear} are all of $O(n+m)$.
The exclusive EMA submodule in \amlptwo~is $O(n)$ as well.
Therefore, the time and memory complexity of AMLP~(both \amlpone~and \amlptwo) is  $O(n+m)$.

\subsubsection{NAR-AMLP}
Non-autoregressive models have one encoder self-attention, one decoder self-attention, and an encoder-decoder cross-attention. 
Due to the quadratic complexity of softmax attention, the complexities of the three attentions are $O(m^2)$, $O(n^2)$ and $O(nm)$, respectively. 
Therefore, the complexity of the entire model architecture is $O(n^2+nm+m^2)$.
To reduce the inefficiency bottlenecked by softmax attention, we replace softmax modules in non-autoregressive models with AMLP, deriving an NAR-AMLP architecture with linear time and space complexity.

\section{Experiments}
\label{all_exp}
We conduct extensive experiments, covering the fields of speech, natural language processing, time-series and computer vision.\footnote{In experiments, we take $\mathrm{softmax}(\cdot)$ as the nonlinear function $\sigma_1(\cdot)$ unless otherwise specified.}
For fair comparison between models, we select the typical hyperparameter setting for each efficient attention on each task, which is shown in Table~\ref{tab:hyperparameters} in detail.
Specifically, we first apply our hybrid architecture NAR-AMLP in two tasks: Text-to-Speech Synthesis and Machine Translation.
Then we assess AMLP's self-attention and cross-attention abilities on super resolution and long sequence time-series forecasting tasks, respectively.
Finally, we conduct ablation studies to show the hidden philosophy of AMLP and explore how efficient AMLP scales to long-sequence modeling.

\begin{table}[tbp]
  \centering
  \caption{Hyperparameters of different tasks. }
  \resizebox{1.0\linewidth}{!}{
    \begin{tabular}{lccccc}
    \toprule
    \textbf{Task} & \multicolumn{2}{c}{TTS} & MT    & SR    & LSTF \\
    \midrule
    \textbf{Backbone} & \multicolumn{2}{c}{\begin{tabular}[c]{@{}c@{}}FastSpeech~2/\\ Transformer-TTS\end{tabular}} & Transformer/CMLMC & SR    & Informer \\
    \midrule
    \midrule
    \multicolumn{6}{c}{\textit{Training hyperparameters}} \\
    \midrule 
    \midrule
    \textbf{Batch Size} & \multicolumn{2}{c}{48} & --     & 4     & 32 \\
    \textbf{Number of Steps (epochs)} & \multicolumn{2}{c}{20K} & 100K/300K & 1M    & 6 (epochs) \\
    \textbf{Warmup Steps} & \multicolumn{2}{c}{4K} & 4K    & --     & -- \\
    \textbf{Peak Learning Rate} & \multicolumn{2}{c}{5e-4} & 5e-4  & 1e-4  & 1e-4 \\
    \textbf{Scheduler} & \multicolumn{2}{c}{Inverse Sqrt} & Inverse Sqrt & Linear & Exponential Decay \\
    \textbf{Optimizer} & \multicolumn{2}{c}{AdamW} & AdamW & AdamW & AdamW \\
    \textbf{Adam} & \multicolumn{2}{c}{(0.9, 0.98)} & (0.9, 0.98) & (0.9, 0.999) & (0.9,0.999) \\
    \textbf{Clip Norm} & \multicolumn{2}{c}{5.0} & 5.0   & 0     & 0 \\
    \textbf{Attention Dropout} & \multicolumn{2}{c}{0.1} & 0.3   & 0.2   & 0.05 \\
    \textbf{Weight Decay} & \multicolumn{2}{c}{0.01} & 0.0001 & 0     & 0 \\
    \textbf{Max Tokens} & \multicolumn{2}{c}{--} & 65536 & --     & -- \\
    \textbf{Iteration} & \multicolumn{2}{c}{--} &   --    & --     & 5 \\
    \textbf{Evaluation Checkpoint} & \multicolumn{2}{c}{best} & average last 10 & average last 5  & last \\
    \midrule
    \midrule
    \multicolumn{6}{c}{\textit{Attention hyperparameters}} \\
    \midrule 
    \midrule
    \textbf{wsize (local)} & \multicolumn{2}{c}{15} & 5 & 15 & 15 \\
    \textbf{landmarks (ABC)} & \multicolumn{2}{c}{64} & 16 & 64 & 64 \\
    \textbf{ffn\_dim (AMLP)} & \multicolumn{2}{c}{64} & 16 & 64 & 64 \\
    \textbf{approx\_dim (Performer)} & \multicolumn{2}{c}{64} & 16 & 64 & 64 \\
    \bottomrule
    \end{tabular}%
    }
  \label{tab:hyperparameters}%
\end{table}%

\subsection{Main Results of NAR-AMLP}

\label{sec:full_linear_experiment}
\subsubsection{Text-to-Speech}
\label{sec:tts_exp}
We select LJSpeech~\cite{ljspeech} dataset for this task, and use FastSpeech~2~(FS2)~\cite{fastspeech2} and Transformer-TTS~(Tr-TTS)~\cite{transformer_tts} as the backbone models for NAR and AR, respectively. 
For both backbones, we replace all softmax attentions with efficient ones to achieve linear complexity.
We use \amlpone~variant and $\mathrm{ReLU}(\cdot)$ as $\sigma_1(\cdot)$ in Eq.~\ref{eq:nonlinear}.
The alignment tool  ``g2pE''~\cite{fairseq_s2s} is applied to train FastSpeech~2.
For reproducibility, we use two widely-used objective evaluation metrics, Mel Cepstral Distortion~(MCD) and Mel Spectral Distortion~(MSD), to assess the quality of synthesized audio clips. 
We compare AMLP with gMLP~\cite{liu2021pay}, XCA~\cite{el2021xcit}, ABC~\cite{abc} and local attention~\cite{luong-etal-2015-effective}. 
The details of training hyperparameters are shown in Table~\ref{tab:hyperparameters}.
We demonstrate the results in Table~\ref{tab:fastspeech_result}.  
AMLP substantially lowers the MCD and MSD values by a great margin up to 0.15 MCD with even lower complexity compared to vanilla models. Additionally, AMLP also outperforms other efficient models. 
Notably, we have significantly lower MCD than XCA which also leverages (cross-)covariance matrices.

\subsubsection{Machine Translation}
\label{sec:mt}
To verify AMLP's capability on short sequence modeling, we launch Machine Translation (MT) experiments on WMT 2014 English-German~(WMT'14 En-De) and German-English~(WMT'14 De-En) datasets~\cite{wmt14dataset}.
We adopt \amlptwo~variant to CMLMC~\cite{huang2022improving}, which is a powerful fully NAR architecture without extra decoding algorithms. 
For completeness, we include widely-used AR architecture Transformer~(Tr)~\cite{vaswani2017attention} with competitive linear attentions. 
We exclude the AR-reranking process to make a fully linear-complexity generation process.
Similar to TTS, we replace self/cross-attention modules in the decoder of Transformer and CMLMC to obtain their efficient variants.
We use hyperparameters as CMLMC and Transformer suggest, which is present in Table ~\ref{tab:hyperparameters}.
We report BLEU-4~\cite{Bleu} scores as the performance metric. 
Because XCA and gMLP do not support cross-attention, we here only compare AMLP with the strong ABC and local baselines.
As translation has implicit token alignment between sequences, local attention can do cross-attention in this task.

Results in Table~\ref{tab:trans_result} indicate that the NAR-AMLP architecture achieves the best result among efficient NAR and AR models with linear complexity.
Among the NAR models, the AMLP model outperforms a strong linear attention model, ABC, on both datasets, with a lead of 0.23 and 0.20 BLEU, respectively. 
It indicates that AMLP effectively captures short-term dependencies and produces more accurate translations than ABC.
We also compare AMLP with vanilla attention, and the results indicate that AMLP outperforms vanilla attention on the de-en dataset, with only a 0.31BLEU lag compared to vanilla attention on the en-de one. 
This suggests that AMLP can achieve comparable performance to vanilla NAR models in certain scenarios.
In comparison to AR models on both datasets, AMLP demonstrates superior performance (with at least 0.22 and 0.24 BLEU improvement), providing further evidence of the efficacy of NAR-AMLP as an architecture.

\begin{figure}[!t]

\begin{minipage}[htbp]{0.45\textwidth}

\makeatletter\def\@captype{table}
  \setlength\aboverulesep{0pt}
  \setlength\belowrulesep{0pt}
  \setlength\cellspacetoplimit{2.5pt}
  \setlength\cellspacebottomlimit{2.5pt}
\caption{Automatic evaluation metric on LJSpeech dataset. All models are trained by ourselves. $n,m$ are the target and source sequence lengths. 
Colored rows represent NAR models.
}
\resizebox{1.0\linewidth}{!}{
    \begin{tabular}{SlSlSlScSc}
    \toprule
    \multirow{1}[5]{*}{\vspace{0.3em}\textbf{Arch}} & \multicolumn{1}{Sl}{\multirow{1}[4]{*}{\textbf{Model}}} & \multicolumn{1}{Sl}{\multirow{1}[4]{*}{\textbf{\#Params}}} & \multicolumn{2}{Sc}{\textbf{LJSpeech}} \\
\cmidrule{4-5}   &       &       & \textbf{MCD}$\downarrow$ & \textbf{MSD}$\downarrow$ \\
    \midrule
    \midrule
    \multicolumn{5}{Sc}{\textit{Complexity: $O(n^2)$ or $O\left(n^2+nm+m^2\right)$}} \\
    \midrule
    \midrule
    AR    & Tr-TTS & 54.40M & 4.095 & 2.199 \\
    \rowcolor[rgb]{ 1,  0.949,  .9} NAR   & FS2 & 41.23M & 3.475 & 1.974 \\
    \midrule
    \midrule
    \multicolumn{5}{Sc}{\textit{Complexity: $O(n)$ or $O(n+m)$}} \\
    \midrule
    \midrule
    AR   & Tr-TTS (ABC) & 54.60M &   5.130     & 2.596 \\
    \rowcolor[rgb]{ 1,  .949,  .9} & FS2 (local)  & 41.23M  & 3.419 &	1.970 \\
    \rowcolor[rgb]{ 1,  .949,  .9}  & FS2 (ABC)  & 41.36M & 3.392 & 1.966 \\
    \rowcolor[rgb]{ 1,  .949,  .9} \multirow{1}[3]{*}{NAR}      & FS2 (XCA)  & 41.23M & 3.500 & 2.024 \\
    \rowcolor[rgb]{ 1,  .949,  .9}       & FS2 (gMLP)  & 44.90M & 3.402 & 1.964 \\
\cmidrule{2-5}    \rowcolor[rgb]{ 1,  .949,  .9}       & FS2 (AMLP) & 41.49M  & \textbf{3.327} & \textbf{1.940} \\
    \bottomrule
    \end{tabular}%
    }
\label{tab:fastspeech_result}

\end{minipage}
\hfill
\begin{minipage}[htbp]{0.45\textwidth}
\vspace{-0.5em}
\makeatletter\def\@captype{table}
  \setlength\aboverulesep{0pt}
  \setlength\belowrulesep{0pt}
  \setlength\cellspacetoplimit{2.5pt}
  \setlength\cellspacebottomlimit{2.5pt}
\caption{BLEU4 scores on WMT14 EN-DE and WMT14 DE-EN dataset. All models for comparison are implemented by ourselves. $n,m$ are the target and source sequence lengths. Colored rows represent NAR models.
}
\resizebox{1.0\linewidth}{!}{

    \begin{tabular}{SlSlSlScSc}
    \toprule
    \multirow{2}[4]{*}{\textbf{Arch}} & \multirow{2}[4]{*}{\textbf{Model}} & \multirow{2}[3]{*}{\textbf{\#Params}} & \multicolumn{2}{Sc}{\textbf{WMT' 14}} \\
\cmidrule{4-5}    &      &       & \boldmath{}\textbf{En-De}\unboldmath{} & \boldmath{}\textbf{De-En}\unboldmath{} \\
    \midrule
    \midrule
    \multicolumn{5}{Sc}{\textit{Complexity: $O\left(n^2+nm+m^2\right)$}} \\
    \midrule
    \midrule
    AR    & Tr & 86.74M  & 27.38 & 31.26 \\
    \rowcolor[rgb]{ 1,  .949,  .898} NAR   & CMLMC  & 73.14M  & \textbf{27.91} & 31.43 \\
    \midrule
    \midrule
    \multicolumn{5}{Sc}{\textit{Complexity: $O(n+m)$}} \\
    \midrule
    \midrule
    \multirow{3}[1]{*}{AR} & Tr (local)  & 86.74M & $\text{24.77}$  & $\text{28.21}$  \\
          & Tr (ABC)   & 86.77M & $\text{25.86}$ & $\text{29.09}$ \\

    \rowcolor[rgb]{ 1,  .949,  .898}       & CMLMC (ABC)  & 73.16M   & $\text{27.37}$ & $\text{31.30}$ \\
    \rowcolor[rgb]{ 1,  .949,  .898}   \multirow{1}[2]{*}{NAR}    & CMLMC (local)  & 73.16M   & $\text{27.05}$ & $\text{30.33}$ \\
    \cmidrule{2-5}
    \rowcolor[rgb]{ 1,  .949,  .898}       & CMLMC (AMLP)  & 73.44M  & $\text{27.60}$ & $\textbf{\text{31.50}}$ \\
    \bottomrule
    \end{tabular}%
    }
\label{tab:trans_result}

\end{minipage}
\end{figure}

\subsection{Self- and Cross-Attention Ablation}
\label{sec:ablate_self_cross}
\subsubsection{Self-attention}
\begin{figure}[!t]

\begin{minipage}[htbp]{0.47\textwidth}

\makeatletter\def\@captype{table}
\caption{PSNR and SSMI on CelebA-HQ dataset. $n$ is the pixel number of the images. }
    \begin{tabular}{llcc}
    \toprule
    \multicolumn{1}{c}{\multirow{2}[2]{*}{\textbf{Model}}} & \multicolumn{1}{c}{\multirow{2}[2]{*}{\textbf{\#Params}}} & \multicolumn{2}{c}{\textbf{Celeb-HQ	
}} \\
      \cmidrule{3-4}             &       & \textbf{PSNR↑} & \textbf{SSMI↑} \\
    \midrule
    \midrule
    \multicolumn{4}{c}{\textit{Complexity:} $O(n^2)$} \\
    \midrule
    \midrule
    vanilla & 99.55M & 23.18 & 0.675 \\
    \midrule
    \midrule
    \multicolumn{4}{c}{\textit{Complexity:} $O(n)$} \\
    \midrule
    \midrule
    local & 99.55M & \textbf{23.33} & 0.682 \\
    gMLP  & 101.66M & 23.24 & 0.679 \\
    XCiT  & 99.55M & 23.08 & 0.67 \\
    ABC   & 99.72M & 22.54 & 0.635 \\
    AMLP  & 99.73M & 23.28 & \textbf{0.684} \\
    \bottomrule
    \end{tabular}%
\label{tab:ablate_sr}

\end{minipage}
\hfill
\begin{minipage}[htbp]{0.47\textwidth}

\makeatletter\def\@captype{figure}

    \caption{Trade-off of MCD value and speed-up of different intermediate dimension $c$ values in text-to-speech task.
    }
    \label{fig:ablate_c}\includegraphics[width=1\linewidth]{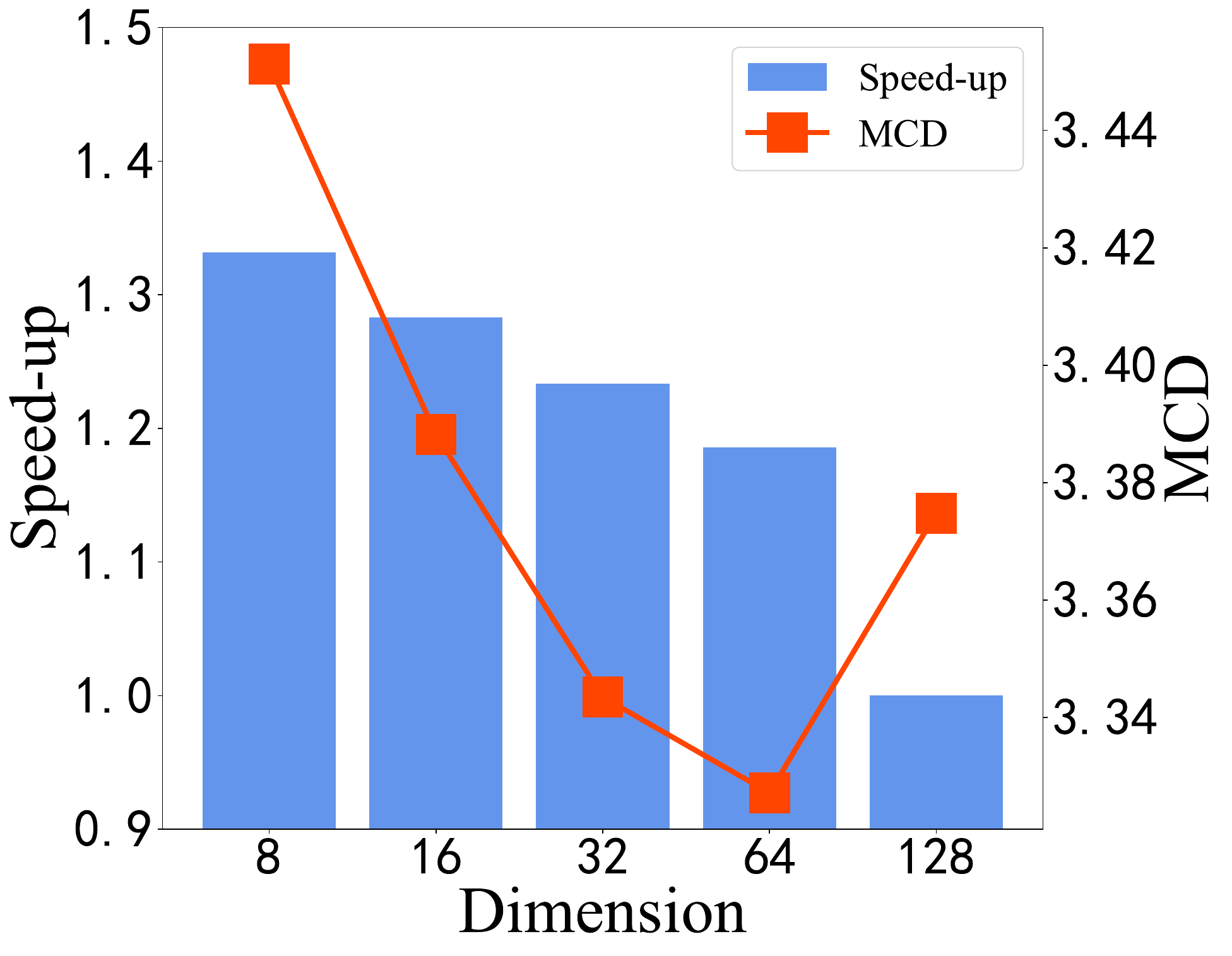}

\end{minipage}
\end{figure}

We evaluate the self-encoding ability of AMLP on Super Resolution~(SR) task.
SR aims to convert low-resolution ($16 \times 16$) images into high-resolution ($128 \times 128$) ones. 
We base on a powerful backbone --- SR3~\cite{sr3} and add attention layers after each residual block to follow CAB~\cite{zhang2022cab} settings. 
We replace the softmax self-attention with five efficient architectures, i.e., local, gMLP, XCA, ABC and AMLP to compare 
Following~\cite{sr3}, we use the Flickr-Faces-HQ (FFHQ) dataset~\cite{ffhq} for the training set and CelebA-HQ dataset~\cite{celebahq} for the evaluation set.
We use Peak Signal-to-Noise Ratio~(PSNR) and Structural SIMilarity~(SSIM)~\cite{wang2004image} to measure efficient models.
Experiment results are shown in Table~\ref{tab:ablate_sr}.
AMLP improves the performance of SR3 to 23.28~(+0.10) on PSNR and 0.684~(+0.09) on SSMI against the vanilla baseline, indicating that AMLP has a strong self-encoding ability.
When compared to gMLP, AMLP also has a slight performance gain.
AMLP outperforms covariance-based architecture XCA by 0.20 and 0.14 on PSNR and SSMI, respectively.
\subsubsection{Cross-Attention}
We test the cross-attention ability on the long sequence time-series forecasting~(LSTF) task. 
We take Informer~\cite{informer} as the backbone neural networks and evaluate efficient models on Electricity Transformer Temperature~(ETT) dataset, which contains three sub-datasets ETT-h1, ETT-h2, and ETT-m1. 
We follow~\cite{informer} to conduct univariate and multivariate evaluations on three sub-datasets and average their Mean Square Error~(MSE) and Mean Absolute Error~(MAE) to obtain final scores.
Except for vanilla attention, we also compare AMLP with other three efficient models with strong cross-alignment abilities: ABC~\cite{abc}, Performer~\cite{choromanski2020rethinking} and cosFormer~\cite{qin2022cosformer}.
We exclude local attention as it does not work for cross attention without explicit token alignment in the time-series forecasting task.
The results performed on three sub-datasets are shown in Table~\ref{tab:ett_full}.
AMLP, in contrast to the vanilla counterpart, achieves lower MSE and MAE as well as more efficient complexity. 
Moreover, we notice that all other efficient models perform poorly compared to vanilla attention. It suggests that AMLP has a solid ability to model non-homologous information.

\begin{table}[tbp]
  \centering
  \small
  \caption{Cross-attention ablation on ETT-h1, ETT-h1, and ETT-m1 datasets. $n,m$ are the target and source lengths. Avg. is computed over three subdatasets.}

    \resizebox{1\linewidth}{!}{
        \begin{tabular}{lllcccccccc}
    \toprule
    \multicolumn{1}{c}{\multirow{2}[4]{*}{\textbf{Complex.}}} & \multicolumn{1}{c}{\multirow{2}[4]{*}{\textbf{Methods}}} & \multicolumn{1}{c}{\multirow{2}[4]{*}{\textbf{\#Params}}} & \multicolumn{2}{c}{\textbf{ETTh1}} & \multicolumn{2}{c}{\textbf{ETTh2}} & \multicolumn{2}{c}{\textbf{ETTm1}} & \multicolumn{2}{c}{\textbf{Avg.}} \\
\cmidrule{4-11}          &       &       & \textbf{MSE↓} & \textbf{MAE↓} & \textbf{MSE↓} & \textbf{MAE↓} & \textbf{MSE↓} & \textbf{MAE↓} & \textbf{MSE↓} & \textbf{MAE↓} \\
    \midrule
    $O(n^2+nm)$ & \textcolor[rgb]{ .122,  .137,  .161}{vanilla} & 11.33M & \textbf{0.754} & \textbf{0.573} & 1.907 & 1.036 & 0.754 & 0.716 & 1.138 & 0.775 \\
    \midrule
    \multirow{4}[2]{*}{$O(n+m)$} & ABC   & 11.33M & 0.845 & 0.728 & 1.862 & 1.013 & 0.734 & 0.685 & 1.147 & 0.809 \\
          & Performer & 11.33M & 0.861 & 0.703 & 2.137 & 1.091 & 0.764 & \textbf{0.663} & 1.254 & 0.819 \\
          & cosFormer & 11.33M & 0.848 & 0.723 & 2.094 & 1.067 & \textbf{0.715} & 0.680 & 1.219 & 0.823 \\
          & AMLP & 11.33M & 0.797 & 0.702 & \textbf{1.504} & \textbf{0.864} & 0.718 & 0.684 & \textbf{1.006} & \textbf{0.750} \\
    \bottomrule
    \end{tabular}%
}
  \label{tab:ett_full}%
\end{table}%

\subsection{Analysis}

In this section, we conduct substantial analysis experiments to dig out the efficiency and superiority of our AMLP mechanism. We first present our analysis in comparison with other efficient attention modules on the TTS task. 
Then we show that our approximation $c<d$ in Eq.~\ref{ea:decom_sigma} does not deteriorate the performance of speech generation. 
Finally, we elucidate the outstanding generation speed and GPU peak usage of our AMLP in the NAR scenario.

\subsubsection{Intermediate Dimension Analysis}
\label{ablate_c}
The approximation of eigenvalues in Eq.~\ref{ea:decom_sigma} prompts us to know whether such approximation is feasible and whether the exorbitant approximation will deteriorate the generation performance. 
To this end, we test several values of $c$ in AMLP and report each corresponding performance on TTS and the decoding speed when adopted to FastSpeech~2, in Fig.~\ref{fig:ablate_c}.
Except for $c$ value, we adopt the same setting in \S\ref{sec:tts_exp}. 

From Fig.~\ref{fig:ablate_c}, we can see that AMLP with approximation rank $c$ can achieve as well as no approximation setting~($c=d=128$) and does not impact the performance greatly.
But with a lower $c$ value, AMLP can achieve better decoding speed. 
% This result for a second time verifies that AMLP approximate the $\mathbf{\Sigma}$ matrix when.
Specifically, in contrast to $c=64$, a higher MCD when setting $c$ to $d$ also indicates that maintaining the whole eigenvalues in Eq.~\ref{ea:decom_sigma} may even lead to overparameterization and impair the overall decoding efficacy.
It verifies the feasibility to approximate $\mathbf{\Sigma}$ with fewer eigenspectrums in AMLP.

\subsubsection{Efficiency Analysis}
\label{efficiency_analysis}

To further understand the performance of NAR-AMLP  architecture in inference, we set up a simulation experiment to test its efficiency.
The simulation experiment evaluates NAR-AMLP efficiency from running time and memory usage with respect to sequence length from 256 to 8,192, compared with AR model and vanilla NAR model.
We simulate the generation process with a single efficient module.
For AR, we test its causal attention, which is its bottleneck in generation.
For AMLP, we use $64$ as the inner dimension with ReLU activation function for $\sigma_1$ in Eq.~\ref{eq:nonlinear}.
\amlpone~and\amlptwo~shares the same complexity, so we use ``AMLP'' to denote the two variants.
The experiments are performed with batch size $12$ on a single A100 GPU, and the results are repeated with 100 runs.
We remain running latency data ranging from the first quatile and the third quatile among the 100 runs to remove noise.
Finally, the remaining figures are averaged to serve as the final time consumption.

\begin{figure*}[tbp]
   
    \begin{subfigure}[c]{0.40\textwidth}
         \centering
         \includegraphics[width=\textwidth]{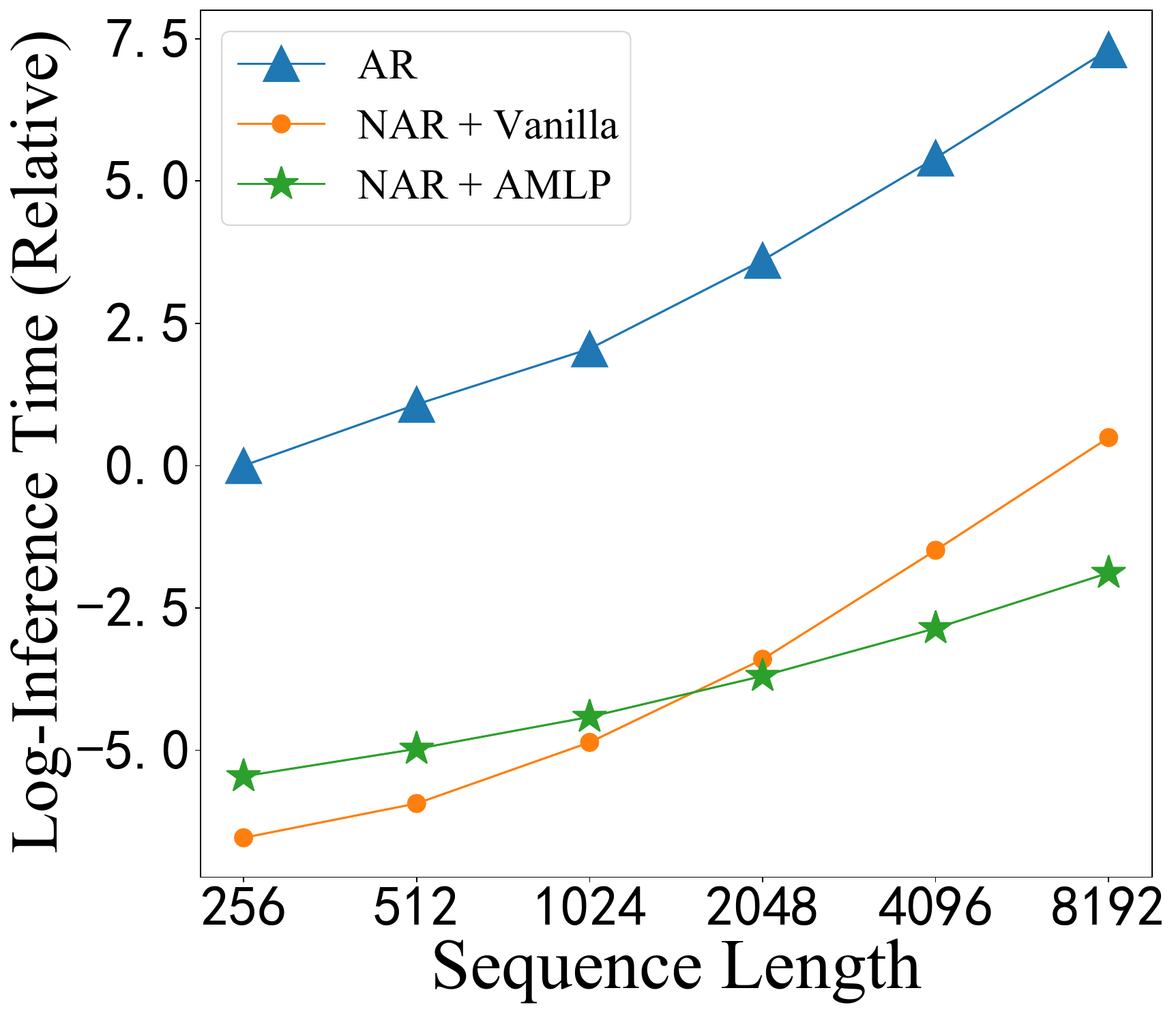}
         \caption{%Empirical test on running time. These results are time consumption of 100 unit tests on average.
         }
         \label{fig:time_compare}
     \end{subfigure}
     \hspace{\fill}
     \begin{subfigure}[c]{0.40\textwidth}
         \centering
         \includegraphics[width=\textwidth]{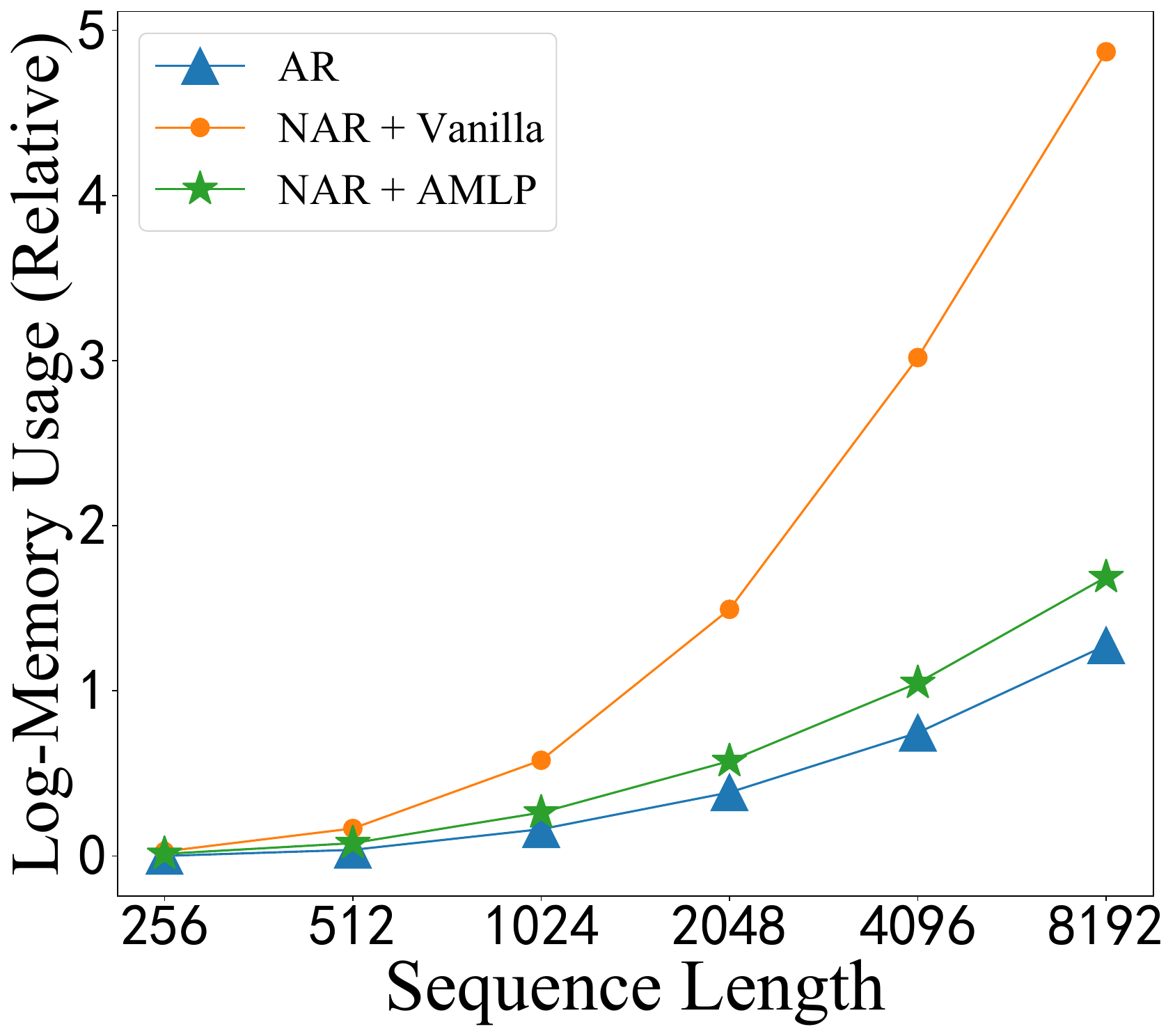}
         \caption{%Empirical test on peak GPU usage. These results are peak memory consumption during the process of computation.
         }
         \label{fig:gpu_compare}
     \end{subfigure}

    \hspace*{\fill}
    \caption{(a)Empirical running time and (b) empirical memory cost with sequence length. Logarithms of relative measurement to the AR model are reported.}
    \label{fig:two_tradeoff}
\end{figure*}

Fig.~\ref{fig:time_compare} shows that NAR-AMLP extremely speeds up the inference process.
To generate a long sequence with $8,192$ tokens, vanilla NAR is $116\times$ faster than AR while NAR-AMLP is even $590\times$ faster.
For sequences with more than $1500$ tokens, both variants of AMLP are more efficient than vanilla attention; otherwise, the vanilla attention is faster.
Fig.~\ref{fig:gpu_compare} shows that NAR-AMLP significantly reduces memory consumption in NAR generation.
It saves $89\%$ memory usage of NAR model when generating a sequence with $8,192$ tokens.
Note that AR models cost fewer memory resources because of incremental decoding, which caches previous states and processes only one token at each step.
But AR models still suffer from huge memory usage as NAR models in training, since they are usually implemented with a causal mask on the attention matrix.
Thus it is reasonable to infer that NAR-AMLP is more efficient than AR and NAR models in training.

\label{ablation}

\section{Related Work}

\subsubsection{Non-Autoregressive Generation}
~\cite{gu2017non} first proposes a non-autoregressive model to generate all the tokens within a sequence in parallel, which extremely speeds up the inference process but is inferior in generation quality. 
To mitigate the quality degradation, many researchers devote to improve the model performance with iterative decoding~\cite{lee2018deterministic,ghazvininejad2019mask,gu2019levenshtein,guo2020jointly,huang2021non}, curriculum learning~\cite{guo2020fine,liu2020task,qian2020glancing,qian2021volctrans,bao2022textit}, latent variable modeling~\cite{ma2019flowseq,ran2019guiding,bao2019non,bao2022textit}, imitation learning~\cite{li2018hint,wei2019imitation} and learning objective~\cite{saharia2020non,ghazvininejad2020aligned,liu2021don,du2021order}.
These previous works focus on pursuing the high efficacy of non-autoregressive generation, but few works are presented to improve NAR's efficiency in long sequence modeling.
We target to further improve its efficiency and scale non-autoregressive models to long sequences.

\subsubsection{MLP Architecture}
Multi-layer perceptron~\cite{gardner1998artificial} is a classic neural network architecture and has been widely used.
Recently, novel variants of MLP architectures are proposed for text and image processing, achieving impressive results on image classification~\cite{tolstikhin2021mlp,liu2021pay}, text classification~\cite{tay2021synthesizer}, multilingual parsing~\cite{fusco2022pnlp}, and intent classification~\cite{fusco2022pnlp}.
MLP-Mixer~\cite{tolstikhin2021mlp} is proposed by leveraging a token-mixing and a channel-mixing MLP to enable token-wise and channel-wise communication.
MLP-Mixer is further improved to pNLP-Mixer with locality sensitive hashing~\cite{indyk1998approximate} projection at the bottom calculating non-trainable fingerprints~\cite{fusco2022pnlp}.
~\cite{liu2021pay} propose gMLP by introducing a spatial gating unit to enhance the communication between neighboring tokens.
CycleMLP~\cite{chen2021cyclemlp} leverages a local window to achieve linear time complexity on dense prediction. 
Besides, previous studies focus on encoding text/image features with MLP, but we explore the possibility to leverage an MLP architecture for sequence generation.

\subsubsection{Attention Mechanism}

Attention is first proposed to align the target and source sequence in neural machine translation~\cite{bahdanau2014neural}, and is further improved to multi-head self/cross/causal attention~\cite{vaswani2017attention}.
Due to its quadratic time complexity and memory cost with sequence length, a surge of efficient attention is proposed to improve the efficiency of softmax attention.
Due to the the sparsity of attention matrix, many researchers propose to explicitly model a sparse attention mechanism to obtain fast computation without harming performance~\cite{ho2019axial,tay2020sparse,kitaev2020reformer,beltagy2020longformer,zaheer2020big,roy2021efficient}.
The low-rank property of attention matrix also brings out matrix decomposition-based methods~\cite{xiong2021nystromformer,chen2021compressed}.
The softmax attention can also be linearized via exponential kernel decomposition~\cite{choromanski2020rethinking,abc,RFA,lara,qin2022cosformer}.
These attention variants are exploring an efficient way to approximate softmax attention, but we focus on MLP architecture, which is naturally an efficient architecture.

\section{Conclusions}
In this work, we introduced Attentive Multi-Layer Perceptron~(AMLP), an efficient plugin alternative to vanilla attention for non-autoregressive generation tasks.
AMLP uses adaptive weights to learn inter-token interactions as done in attention.
And we also put forward two methods adopting different philosophies to parameterize the adaptive weight matrices in AMLP.
Substantial experiments on generation tasks verify that AMLP surpasses attention in most tasks and achieves similar performances with other strong efficient models in other tasks. 
Besides, efficiency analysis indicates that AMLP combined NAR model could save time compared to AR models, and save space compared to vanilla NAR models in long sequence settings.

\section{Ethical Issues}
AMLP is designed to speed up the generation of non-autoregressive models, by replacing the inefficient softmax attention with our AMLP module to achieve linear complexity.
The potential positive implications imply lower difficulty in deploying NAR models on resource-limited devices, thus increasing the accessibility of NAR models.
AMLP also makes positive impacts on extending NAR models to various domains, since it can do both self-attention and cross-attention.
Moreover, the high efficiency of AMLP reduces the carbon footprint of training a model and thus brings positive environmental benefits.
As such, we do not foresee any immediate negative ethical or societal consequences stemming from our work that are different from those that apply to other fundamental components of the transformer architecture and NAR models.

\bibliographystyle{splncs04}
\bibliography{mybibliography}

\begin{thebibliography}{10}
\providecommand{\url}[1]{\texttt{#1}}
\providecommand{\urlprefix}{URL }
\providecommand{\doi}[1]{https://doi.org/#1}

\bibitem{el2021xcit}
Ali, A., Touvron, H., Caron, M., Bojanowski, P., Douze, M., Joulin, A., Laptev, I., Neverova, N., Synnaeve, G., Verbeek, J., et~al.: Xcit: Cross-covariance image transformers. Advances in neural information processing systems  \textbf{34},  20014--20027 (2021)

\bibitem{bahdanau2014neural}
Bahdanau, D., Cho, K., Bengio, Y.: Neural machine translation by jointly learning to align and translate. In: ICLR (2015)

\bibitem{bao2019non}
Bao, Y., Zhou, H., Feng, J., Wang, M., Huang, S., Chen, J., Li, L.: Non-autoregressive transformer by position learning. arXiv preprint arXiv:1911.10677  (2019)

\bibitem{bao2022textit}
Bao, Y., Zhou, H., Huang, S., Wang, D., Qian, L., Dai, X., Chen, J., Li, L.: {latent-GLAT}: Glancing at latent variables for parallel text generation. In: Proceedings of the 60th Annual Meeting of the Association for Computational Linguistics (Volume 1: Long Papers). pp. 8398--8409. Association for Computational Linguistics, Dublin, Ireland (May 2022). \doi{10.18653/v1/2022.acl-long.575}, \url{https://aclanthology.org/2022.acl-long.575}

\bibitem{beltagy2020longformer}
Beltagy, I., Peters, M.E., Cohan, A.: Longformer: The long-document transformer. arXiv preprint arXiv:2004.05150  (2020)

\bibitem{wmt14dataset}
Bojar, O., Buck, C., Federmann, C., Haddow, B., Koehn, P., Leveling, J., Monz, C., Pecina, P., Post, M., Saint-Amand, H., Soricut, R., Specia, L., Tamchyna, A.: Findings of the 2014 workshop on statistical machine translation. In: Proceedings of the Ninth Workshop on Statistical Machine Translation. pp. 12--58. Association for Computational Linguistics, Baltimore, Maryland, USA (Jun 2014). \doi{10.3115/v1/W14-3302}

\bibitem{chang2022maskgit}
Chang, H., Zhang, H., Jiang, L., Liu, C., Freeman, W.T.: Maskgit: Masked generative image transformer. In: Proceedings of the IEEE/CVF Conference on Computer Vision and Pattern Recognition. pp. 11315--11325 (2022)

\bibitem{chen2021cyclemlp}
Chen, S., Xie, E., GE, C., Chen, R., Liang, D., Luo, P.: Cycle{MLP}: A {MLP}-like architecture for dense prediction. In: International Conference on Learning Representations (2022), \url{https://openreview.net/forum?id=NMEceG4v69Y}

\bibitem{chen2021compressed}
Chen, Z., Gong, M., Ge, L., Du, B.: Compressed self-attention for deep metric learning with low-rank approximation. In: Proceedings of the Twenty-Ninth International Conference on International Joint Conferences on Artificial Intelligence. pp. 2058--2064 (2021)

\bibitem{choromanski2020rethinking}
Choromanski, K.M., Likhosherstov, V., Dohan, D., Song, X., Gane, A., Sarlos, T., Hawkins, P., Davis, J.Q., Mohiuddin, A., Kaiser, L., Belanger, D.B., Colwell, L.J., Weller, A.: Rethinking attention with performers. In: International Conference on Learning Representations (2021), \url{https://openreview.net/forum?id=Ua6zuk0WRH}

\bibitem{dosovitskiy2020image}
Dosovitskiy, A., Beyer, L., Kolesnikov, A., Weissenborn, D., Zhai, X., Unterthiner, T., Dehghani, M., Minderer, M., Heigold, G., Gelly, S., Uszkoreit, J., Houlsby, N.: An image is worth 16x16 words: Transformers for image recognition at scale. In: International Conference on Learning Representations (2021)

\bibitem{du2021order}
Du, C., Tu, Z., Jiang, J.: Order-agnostic cross entropy for non-autoregressive machine translation. In: International Conference on Machine Learning. pp. 2849--2859. PMLR (2021)

\bibitem{fusco2022pnlp}
Fusco, F., Pascual, D., Staar, P.: pnlp-mixer: an efficient all-mlp architecture for language. arXiv preprint arXiv:2202.04350  (2022)

\bibitem{gardner1998artificial}
Gardner, M.W., Dorling, S.: Artificial neural networks (the multilayer perceptron)—a review of applications in the atmospheric sciences. Atmospheric environment  \textbf{32}(14-15),  2627--2636 (1998)

\bibitem{ghazvininejad2020aligned}
Ghazvininejad, M., Karpukhin, V., Zettlemoyer, L., Levy, O.: Aligned cross entropy for non-autoregressive machine translation. In: International Conference on Machine Learning. pp. 3515--3523. PMLR (2020)

\bibitem{ghazvininejad2019mask}
Ghazvininejad, M., Levy, O., Liu, Y., Zettlemoyer, L.: Mask-predict: Parallel decoding of conditional masked language models. In: Proceedings of the 2019 Conference on Empirical Methods in Natural Language Processing and the 9th International Joint Conference on Natural Language Processing (EMNLP-IJCNLP). pp. 6112--6121 (2019)

\bibitem{gu2017non}
Gu, J., Bradbury, J., Xiong, C., Li, V.O., Socher, R.: Non-autoregressive neural machine translation. In: International Conference on Learning Representations (2018)

\bibitem{gu2019levenshtein}
Gu, J., Wang, C., Zhao, J.: Levenshtein transformer. Advances in Neural Information Processing Systems  \textbf{32} (2019)

\bibitem{guo2020fine}
Guo, J., Tan, X., Xu, L., Qin, T., Chen, E., Liu, T.Y.: Fine-tuning by curriculum learning for non-autoregressive neural machine translation. Proceedings of the AAAI Conference on Artificial Intelligence  \textbf{34}(05),  7839--7846 (Apr 2020). \doi{10.1609/aaai.v34i05.6289}, \url{https://ojs.aaai.org/index.php/AAAI/article/view/6289}

\bibitem{guo2020jointly}
Guo, J., Xu, L., Chen, E.: Jointly masked sequence-to-sequence model for non-autoregressive neural machine translation. In: Proceedings of the 58th Annual Meeting of the Association for Computational Linguistics. pp. 376--385 (2020)

\bibitem{ho2019axial}
Ho, J., Kalchbrenner, N., Weissenborn, D., Salimans, T.: Axial attention in multidimensional transformers. arXiv preprint arXiv:1912.12180  (2019)

\bibitem{huang2021non}
Huang, C., Zhou, H., Za{\"\i}ane, O.R., Mou, L., Li, L.: Non-autoregressive translation with layer-wise prediction and deep supervision. In: Proceedings of the AAAI Conference on Artificial Intelligence. vol.~36, pp. 10776--10784 (2022)

\bibitem{huang2022improving}
Huang, X.S., Perez, F., Volkovs, M.: Improving non-autoregressive translation models without distillation. In: International Conference on Learning Representations (2022)

\bibitem{indyk1998approximate}
Indyk, P., Motwani, R.: Approximate nearest neighbors: towards removing the curse of dimensionality. In: Proceedings of the thirtieth annual ACM symposium on Theory of computing. pp. 604--613 (1998)

\bibitem{ljspeech}
Ito, K., Johnson, L.: The lj speech dataset. \url{https://keithito.com/LJ-Speech-Dataset/} (2017)

\bibitem{celebahq}
Karras, T., Aila, T., Laine, S., Lehtinen, J.: Progressive growing of gans for improved quality, stability, and variation. In: International Conference on Learning Representations (2018)

\bibitem{ffhq}
Karras, T., Laine, S., Aila, T.: A style-based generator architecture for generative adversarial networks. In: Proceedings of the IEEE/CVF Conference on Computer Vision and Pattern Recognition (CVPR) (June 2019)

\bibitem{kitaev2020reformer}
Kitaev, N., Kaiser, L., Levskaya, A.: Reformer: The efficient transformer. In: International Conference on Learning Representations (2020), \url{https://openreview.net/forum?id=rkgNKkHtvB}

\bibitem{lee2018deterministic}
Lee, J., Mansimov, E., Cho, K.: Deterministic non-autoregressive neural sequence modeling by iterative refinement. In: Proceedings of the 2018 Conference on Empirical Methods in Natural Language Processing. pp. 1173--1182 (2018)

\bibitem{transformer_tts}
Li, N., Liu, S., Liu, Y., Zhao, S., Liu, M.: Neural speech synthesis with transformer network. Proceedings of the AAAI Conference on Artificial Intelligence  \textbf{33}(01),  6706--6713 (Jul 2019). \doi{10.1609/aaai.v33i01.33016706}, \url{https://ojs.aaai.org/index.php/AAAI/article/view/4642}

\bibitem{li2018hint}
Li, Z., Lin, Z., He, D., Tian, F., Qin, T., Wang, L., Liu, T.Y.: Hint-based training for non-autoregressive machine translation. In: Proceedings of the 2019 Conference on Empirical Methods in Natural Language Processing and the 9th International Joint Conference on Natural Language Processing (EMNLP-IJCNLP). pp. 5708--5713. Association for Computational Linguistics, Hong Kong, China (Nov 2019). \doi{10.18653/v1/D19-1573}, \url{https://aclanthology.org/D19-1573}

\bibitem{liu2021don}
Liu, G., Yang, Z., Tao, T., Liang, X., Bao, J., Li, Z., He, X., Cui, S., Hu, Z.: Don{'}t take it literally: An edit-invariant sequence loss for text generation. In: Proceedings of the 2022 Conference of the North American Chapter of the Association for Computational Linguistics: Human Language Technologies. pp. 2055--2078. Association for Computational Linguistics, Seattle, United States (Jul 2022). \doi{10.18653/v1/2022.naacl-main.150}, \url{https://aclanthology.org/2022.naacl-main.150}

\bibitem{liu2021pay}
Liu, H., Dai, Z., So, D., Le, Q.V.: Pay attention to mlps. Advances in Neural Information Processing Systems  \textbf{34},  9204--9215 (2021)

\bibitem{liu2020task}
Liu, J., Ren, Y., Tan, X., Zhang, C., Qin, T., Zhao, Z., Liu, T.Y.: Task-level curriculum learning for non-autoregressive neural machine translation. In: IJCAI. pp. 3861--3867 (2020)

\bibitem{liu2021swin}
Liu, Z., Lin, Y., Cao, Y., Hu, H., Wei, Y., Zhang, Z., Lin, S., Guo, B.: Swin transformer: Hierarchical vision transformer using shifted windows. In: Proceedings of the IEEE/CVF International Conference on Computer Vision. pp. 10012--10022 (2021)

\bibitem{luong-etal-2015-effective}
Luong, T., Pham, H., Manning, C.D.: Effective approaches to attention-based neural machine translation. In: Proceedings of the 2015 Conference on Empirical Methods in Natural Language Processing. pp. 1412--1421. Association for Computational Linguistics, Lisbon, Portugal (Sep 2015). \doi{10.18653/v1/D15-1166}, \url{https://aclanthology.org/D15-1166}

\bibitem{ma2022mega}
Ma, X., Zhou, C., Kong, X., He, J., Gui, L., Neubig, G., May, J., Zettlemoyer, L.: Mega: moving average equipped gated attention. arXiv preprint arXiv:2209.10655  (2022)

\bibitem{ma2019flowseq}
Ma, X., Zhou, C., Li, X., Neubig, G., Hovy, E.: {F}low{S}eq: Non-autoregressive conditional sequence generation with generative flow. In: Proceedings of the 2019 Conference on Empirical Methods in Natural Language Processing and the 9th International Joint Conference on Natural Language Processing (EMNLP-IJCNLP). pp. 4282--4292. Association for Computational Linguistics, Hong Kong, China (Nov 2019). \doi{10.18653/v1/D19-1437}

\bibitem{Bleu}
Papineni, K., Roukos, S., Ward, T., Zhu, W.J.: Bleu: a method for automatic evaluation of machine translation. In: Proceedings of the 40th annual meeting of the Association for Computational Linguistics. pp. 311--318 (2002)

\bibitem{abc}
Peng, H., Kasai, J., Pappas, N., Yogatama, D., Wu, Z., Kong, L., Schwartz, R., Smith, N.A.: {ABC}: Attention with bounded-memory control. In: Proceedings of the 60th Annual Meeting of the Association for Computational Linguistics (Volume 1: Long Papers). pp. 7469--7483. Association for Computational Linguistics, Dublin, Ireland (May 2022). \doi{10.18653/v1/2022.acl-long.515}, \url{https://aclanthology.org/2022.acl-long.515}

\bibitem{RFA}
Peng, H., Pappas, N., Yogatama, D., Schwartz, R., Smith, N., Kong, L.: Random feature attention. In: International Conference on Learning Representations (2021)

\bibitem{qian2020glancing}
Qian, L., Zhou, H., Bao, Y., Wang, M., Qiu, L., Zhang, W., Yu, Y., Li, L.: Glancing transformer for non-autoregressive neural machine translation. In: Proceedings of the 59th Annual Meeting of the Association for Computational Linguistics and the 11th International Joint Conference on Natural Language Processing (Volume 1: Long Papers). pp. 1993--2003. Association for Computational Linguistics, Online (Aug 2021). \doi{10.18653/v1/2021.acl-long.155}

\bibitem{qian2021volctrans}
Qian, L., Zhou, Y., Zheng, Z., Zhu, Y., Lin, Z., Feng, J., Cheng, S., Li, L., Wang, M., Zhou, H.: The volctrans {GLAT} system: Non-autoregressive translation meets {WMT}21. In: Proceedings of the Sixth Conference on Machine Translation. pp. 187--196. Association for Computational Linguistics, Online (Nov 2021)

\bibitem{qin2022cosformer}
Qin, Z., Sun, W., Deng, H., Li, D., Wei, Y., Lv, B., Yan, J., Kong, L., Zhong, Y.: cosformer: Rethinking softmax in attention. In: International Conference on Learning Representations (2022), \url{https://openreview.net/forum?id=Bl8CQrx2Up4}

\bibitem{ran2019guiding}
Ran, Q., Lin, Y., Li, P., Zhou, J.: Guiding non-autoregressive neural machine translation decoding with reordering information. In: AAAI. pp. 13727--13735 (2021)

\bibitem{fastspeech2}
Ren, Y., Hu, C., Tan, X., Qin, T., Zhao, S., Zhao, Z., Liu, T.Y.: Fastspeech 2: Fast and high-quality end-to-end text to speech. In: International Conference on Learning Representations (2021)

\bibitem{roy2021efficient}
Roy, A., Saffar, M., Vaswani, A., Grangier, D.: Efficient content-based sparse attention with routing transformers. Transactions of the Association for Computational Linguistics  \textbf{9},  53--68 (2021)

\bibitem{saharia2020non}
Saharia, C., Chan, W., Saxena, S., Norouzi, M.: Non-autoregressive machine translation with latent alignments. In: Proceedings of the 2020 Conference on Empirical Methods in Natural Language Processing (EMNLP). pp. 1098--1108. Association for Computational Linguistics, Online (Nov 2020). \doi{10.18653/v1/2020.emnlp-main.83}

\bibitem{sr3}
Saharia, C., Ho, J., Chan, W., Salimans, T., Fleet, D.J., Norouzi, M.: Image super-resolution via iterative refinement. IEEE Transactions on Pattern Analysis and Machine Intelligence  (2022)

\bibitem{tay2021synthesizer}
Tay, Y., Bahri, D., Metzler, D., Juan, D.C., Zhao, Z., Zheng, C.: Synthesizer: Rethinking self-attention for transformer models. In: International Conference on Machine Learning. pp. 10183--10192. PMLR (2021)

\bibitem{tay2020sparse}
Tay, Y., Bahri, D., Yang, L., Metzler, D., Juan, D.C.: Sparse sinkhorn attention. In: International Conference on Machine Learning. pp. 9438--9447. PMLR (2020)

\bibitem{tolstikhin2021mlp}
Tolstikhin, I.O., Houlsby, N., Kolesnikov, A., Beyer, L., Zhai, X., Unterthiner, T., Yung, J., Steiner, A., Keysers, D., Uszkoreit, J., et~al.: Mlp-mixer: An all-mlp architecture for vision. Advances in Neural Information Processing Systems  \textbf{34} (2021)

\bibitem{vaswani2017attention}
Vaswani, A., Shazeer, N., Parmar, N., Uszkoreit, J., Jones, L., Gomez, A.N., Kaiser, {\L}., Polosukhin, I.: Attention is all you need. Advances in neural information processing systems  \textbf{30} (2017)

\bibitem{fairseq_s2s}
Wang, C., Hsu, W.N., Adi, Y., Polyak, A., Lee, A., Chen, P.J., Gu, J., Pino, J.: fairseq s{\^{}}2: A scalable and integrable speech synthesis toolkit. In: Proceedings of the 2021 Conference on Empirical Methods in Natural Language Processing: System Demonstrations. pp. 143--152. Association for Computational Linguistics, Online and Punta Cana, Dominican Republic (Nov 2021). \doi{10.18653/v1/2021.emnlp-demo.17}

\bibitem{wang2004image}
Wang, Z., Bovik, A.C., Sheikh, H.R., Simoncelli, E.P.: Image quality assessment: from error visibility to structural similarity. IEEE transactions on image processing  \textbf{13}(4),  600--612 (2004)

\bibitem{wei2019imitation}
Wei, B., Wang, M., Zhou, H., Lin, J., Sun, X.: Imitation learning for non-autoregressive neural machine translation. In: Proceedings of the 57th Annual Meeting of the Association for Computational Linguistics. pp. 1304--1312. Association for Computational Linguistics, Florence, Italy (Jul 2019). \doi{10.18653/v1/P19-1125}

\bibitem{xiong2021nystromformer}
Xiong, Y., Zeng, Z., Chakraborty, R., Tan, M., Fung, G., Li, Y., Singh, V.: Nyströmformer: A nyström-based algorithm for approximating self-attention. Proceedings of the AAAI Conference on Artificial Intelligence  \textbf{35}(16),  14138--14148 (May 2021). \doi{10.1609/aaai.v35i16.17664}, \url{https://ojs.aaai.org/index.php/AAAI/article/view/17664}

\bibitem{zaheer2020big}
Zaheer, M., Guruganesh, G., Dubey, K.A., Ainslie, J., Alberti, C., Ontanon, S., Pham, P., Ravula, A., Wang, Q., Yang, L., et~al.: Big bird: Transformers for longer sequences. Advances in Neural Information Processing Systems  \textbf{33},  17283--17297 (2020)

\bibitem{zhang2022cab}
Zhang, J., Jiang, S., Feng, J., Zheng, L., Kong, L.: Cab: Comprehensive attention benchmarking on long sequence modeling. arXiv preprint arXiv:2210.07661  (2022)

\bibitem{lara}
Zheng, L., Wang, C., Kong, L.: Linear complexity randomized self-attention mechanism. arXiv preprint arXiv:2204.04667  (2022)

\bibitem{informer}
Zhou, H., Zhang, S., Peng, J., Zhang, S., Li, J., Xiong, H., Zhang, W.: Informer: Beyond efficient transformer for long sequence time-series forecasting. Proceedings of the AAAI Conference on Artificial Intelligence  \textbf{35}(12),  11106--11115 (May 2021), \url{https://ojs.aaai.org/index.php/AAAI/article/view/17325}

\end{thebibliography}





% LNCS Homepage, \url{http://www.springer.com/lncs}. Last accessed 4
% \end{thebibliography}
\end{document}